\pdfoutput=1

\documentclass[11pt]{article}

\usepackage[preprint]{acl}

\usepackage{times}
\usepackage{latexsym}

\usepackage[T1]{fontenc}

\usepackage[utf8]{inputenc}

\usepackage{microtype}

\usepackage{inconsolata}

\usepackage{graphicx}
\usepackage{hyperref}
\usepackage{booktabs}
\usepackage{wasysym}
\usepackage{tabularx}
\usepackage{array}
\usepackage{enumitem}
\usepackage{forest}
\usepackage{longtable}

\newcommand{\spacerow}{\addlinespace[0.7em]}

\title{Web(er) of Hate: A Survey on How Hate Speech Is Typed}

\author{Luna Wang \and Andrew Caines \and Alice Hutchings \\
        Department of Computer Science \& Technology \\ University of Cambridge \\ Cambridge, UK\\
        \texttt{\{cw829, apc38, ah793\}@cam.ac.uk}
 }

\begin{document}
\maketitle
\begin{abstract}
The curation of hate speech datasets involves complex design decisions that balance competing priorities. This paper critically examines these methodological choices in a diverse range of datasets, highlighting common themes and practices, and their implications for dataset reliability. Drawing on Max Weber’s notion of ideal types, we argue for a reflexive approach in dataset creation, urging researchers to acknowledge their own value judgments during dataset construction, fostering transparency and methodological rigour.

Warning:
This document contains examples of hateful content in Section 6.
\end{abstract}

\section{Introduction}

Researchers in computer science, particularly within the NLP community, are increasingly devoting attention to online hate speech. 
As a deeply social phenomenon, \textit{online} hate speech has been recognised in prior research for its potential to incite and propagate \textit{offline} violence \cite{Lupu_2023_offline}. Since \citet{waseem-hovy-2016-hateful-fixed}, there have been a plethora of hate speech datasets\footnote{In this paper, we use the term ``hate speech dataset'' in its widest sense. We include datasets covering hate speech, abusive language, offensive language, and to a lesser extent harassment and cyberbullying as well as other types of text-based online harms, as described by their corresponding authors.} with great diversity in their curation processes despite sharing the overarching goal of advancing state-of-the-art hate speech detection.
As noted by previous research, this heterogeneity negatively affects cross-dataset and cross-domain generalisation \cite{Yin_Zubiaga_2021, Guimaraes_2023_anatomy}. At the same time, it has opened up other research directions, such as transfer learning \cite{Ali_2022_hate}.

While the differences in datasets are highlighted in past survey studies \cite{Fortuna_2019_survey, Poletto_Basile_Sanguinetti_Bosco_Patti_2021}, areas such as design goal and quality assurance are often overlooked. 
In this paper, we draw on Max Weber's notion of ``ideal types'' \cite{weber1904objectivity, weber1930protestant, weber1978economy} (see \S\ref{sec:two}) to highlight that the diversity in hate speech datasets are natural and unavoidable. Instead of pursuing definitional completeness, researchers should adopt a reflexive dataset curation approach. We argue that a fully accurate and comprehensive decomposition of hate speech might not exist. Instead, to progress as a field, the complexities of hate speech should be recognised and the perspectives and assumptions of researchers documented.


We aim to answer the following research question: 
\textit{After deciding to curate a labelled corpus for hate speech detection, how has past research defined hate speech and how do the design decisions differ?}
In doing so, we make the following contributions:
\begin{itemize}
    \item We apply Weber’s ideal types of social action to hate speech datasets, offering a structured framework for understanding socio-political drivers behind hate speech.
    \item We propose a reflexive approach to dataset curation, encouraging researchers to critically examine and document value judgments and frames of reference to promote transparency.
    \item We highlight the impact of annotator composition, contrasting smaller, curated annotator pools suited for prescriptive guidelines with more diverse, crowdsourced datasets better aligned with descriptive approaches.
    \item We critique annotation aggregation practices, advocating alternative ways to capture diverse perspectives and avoid oversimplification.
\end{itemize}

We provide an overview of Weber's ideal types~(\S\ref{sec:two}) and previous surveys (\S\ref{sec_related}). Paper selection is outlined in~\S\ref{sec_selection}. In~\S\ref{sec_findings}, we outline key insights and observations. Our discussion (\S\ref{sec_discussion}) synthesises and interprets our findings. The Appendix includes breakdowns of the datasets analysed.

\section{Weber's Ideal Types}
\label{sec:two}

The inherent subjectivity and the variability in defining hate speech have been discussed within the NLP community \cite{Fortuna_2019_survey, vidgen-direction-2021, pachinger-etal-2023-toward}. This subjectiveness makes hate speech detection as a classification task difficult. 
In discussing the subjectivity of hate speech detection, \citet{rottger-etal-2022-two} outline two contrasting paradigms to encourage researchers to either embrace or limit the subjectivity of the task to the fullest extent. \citet{cercas-curry-etal-2024-subjective-fixed} call for a separation between \textit{-ism}s and offence and distinguish individual differences from subjectivity.

\textit{Ideal types}, conceived by the German sociologist Max Weber, are analytical heuristics that serve to make sense of complex social phenomena. They are not perfectly all-encompassing, nor do they represent the average. Rather, in an observer's attempts to understand phenomena such as capitalism \cite{weber1930protestant} or, more relevant to this discussion, hate speech, these \textit{ideal} 
constructs are created to ``sort out'' the underlying complexities. It is therefore inevitable that these constructs depend on the observer's frame of reference, and as a result the observer---whether consciously or unconsciously---articulates certain aspects that they deem worthy while suppressing those of less importance. 

Viewed through a Weberian lens, the subjectivity and variation of hate speech datasets are grounded in the frame of reference (cultural norms, historical perspectives, laws, moderation guidelines, and values) that actors (researchers from computer science, linguistics, gender/political/religious studies, criminology or law, annotators, platforms, moderators, speakers, recipients, bystanders, and counterspeech campaigners) choose to adopt and accept. 
Prescriptive guidelines can limit variation \cite{rottger-etal-2022-two}, but may still introduce bias through the identity and values of the moderator, speaker, and recipient.

Weber names four ideal types of social action:\footnote{As they are ideal types, they are not mutually exclusive and real world examples often exhibit properties of multiple types at the same time.}
    
\noindent\textbf{Goal-rational} (\textit{zweckrational}): motivated by precise and strategic calculation with the aim of achieving some goals.

\noindent\textbf{Value-rational} (\textit{wertrational}): motivated by values and beliefs despite their potentially sub-optimal consequences.

\noindent \textbf{Affectual} (\textit{affektuell}): driven by emotions.

\noindent \textbf{Traditional} (\textit{traditional}): based on established traditions and habits.

In the context of hate speech, \textbf{goal-rationality} might see hate speech being used strategically to achieve political or ideological goals. Researchers might be interested in how such discourse polarises public opinions and even radicalises the public to the extremes.
From a \textbf{value-rational} perspective, hate speech might be expressed in ways that align with the speaker's beliefs about race, gender, or religion. The evaluation of such belief-driven hate speech is heavily dependent on whether the observer (e.g.\ a researcher, moderator, annotator, or a set of annotation guidelines) shares those values.
\textbf{Affectual action} hate speech can be an emotional response, such as anger or frustration. This category is relevant when considering hate speech in interpersonal conflicts such as Wikipedia or code repository edit comments. Moderators might struggle with distinguishing these reactionary expressions of emotions from more systematic hate speech. 
Finally, \textbf{traditional} forms of hate speech are embedded in cultural and societal norms and traditions, such as casual misogyny or transphobia in some communities. This, too, requires the observer to be aware of their tradition and how it might affect their judgement of hate.

By operationalising their concept of hate speech, researchers risk missing aspects of discourse that do not fit neatly with their ideal type. For example, anti-Semitic conspiracy theories often do not contain explicit slurs but rely on coded language and misinformation (e.g.\ accusations of global control)~\cite{rathje}. These types of covert, goal-driven hate have been overlooked by previous ideal types of hate speech. At the same time, however, it is unrealistic and perhaps impossible to create a perfect representation of hate speech. Researchers must rely on using ideal types to study the areas in focus, and any ideal type is an idealised representation, bound to overlook certain aspects.

Actors use frames of reference to construct an ideal type. Goal-rational actions, such as online moderation, may use prescribed guidelines. However, these are not stable, and the terms of reference can change over time and place. Meta and X (formerly Twitter) have changed their policies regarding transphobic hate speech. This highlights the challenge of developing prescriptive guidelines that remain relevant and applicable.

By recognising that any operationalisation of hate speech is an ideal-typical construct, we argue no single decomposition can fully encapsulate the complexity of hate speech. Instead, researchers should explicitly document their perspectives and assumptions, acknowledging the underlying subjectivities in their operationalisation.


\section{Related Work}
\label{sec_related}
\citet{Poletto_Basile_Sanguinetti_Bosco_Patti_2021} provide the most comparable survey of hate speech datasets, reviewing 64 datasets across five dimensions. In contrast, our study doubles the coverage, making it the most comprehensive to date, but adopts a distinct stance on operationalisation. While \citet{Poletto_Basile_Sanguinetti_Bosco_Patti_2021} advocate for shared operational frameworks and benchmark resources, we draw on Weberian theory to argue that frameworks and evaluations should be tailored to datasets and models individually in relation to their specific purpose and the curator's ideal-typical operationalisation.

\citet{Yu_2024_unseen} review 492 datasets, focussing on the targeted identities within hate speech datasets and revealing discrepancies between conceptualised, operationalised, and detected targets, leading to inconsistencies in hate speech classification models.
\citet{tonneau-etal-2024-languages} review 75 hate speech datasets across languages and geo-cultural contexts, revealing a diminishing English-language bias but persistent over-representation of countries like the US and UK. 

While their work provides valuable insights into identity and geo-cultural representation, our study takes a broader approach by examining the entire dataset curation process, including definitions, intended goals, and design choices. The biases revealed by \citet{Yu_2024_unseen} and \citet{tonneau-etal-2024-languages} illustrate the gap between curators' ideal types---as conceptualised in their definitions and frameworks---and the realities of their final datasets, reinforcing our argument that dataset validity hinges on alignment with intended objectives rather than definitional completeness.

\section{Selection Criteria}
\label{sec_selection}
The primary source of our datasets is the community-maintained Hate Speech Dataset Catalogue\footnote{\url{hatespeechdata.com}} \cite{vidgen-direction-2021}, which lists 124 research papers and their associated datasets across 25 languages but has limited coverage post-2023. To supplement this, we conducted a Google Scholar search paying particular attention to two venues. 
Specifically, we conducted two targeted searches and one general search using the following query: 
\begin{quote}
     (“hate” OR “hates” OR “hateful” OR “offensive” OR “offence” OR “offensiveness” OR “harass” OR “harassing” OR “harassment” OR "aggressive" OR "aggressiveness") AND "dataset".
\end{quote}
We chose these keywords to broadly cover terms commonly used in existing literature. While we acknowledge scope-specific keywords such as ``racism'' and ``sexism'', we did not include those to avoid biasing the search towards specific types of hate. 

To target ACL (Association for Computational Linguistics) and ACM (Association for Computing Machinery), we suffix \texttt{site:aclanthology.org} and \texttt{site:acm.org} to the query respectively. For general search, we append their negative filters to reduce redundancy. 

We filter results to studies published from 2023 onward, considering only the first three pages of search results. We only select studies that introduce and describe a new dataset. Non-textual-content-based prediction \citep[e.g.\ predicting using metadata, ][] {Casavantes-Leveraging-2023} are excluded, but re-labelled datasets are included along with their originals.\footnote{The ACL, ACM, and general searches were conducted on 25 Jan 2025, 3 Feb 2025, and 9 Feb 2025 respectively.} 
We verify consistency across multiple top-level domains (\texttt{.com}, \texttt{.co.uk}, \texttt{.jp}, and \texttt{.hk}). The search is conducted in incognito mode to remove any potential search engine personalisation. We do not conduct a full snowballing process due to its bias toward older studies and limited added value beyond our combined search strategy. 

\begin{figure}[ht]
    \centering
    \includegraphics[width=\linewidth]{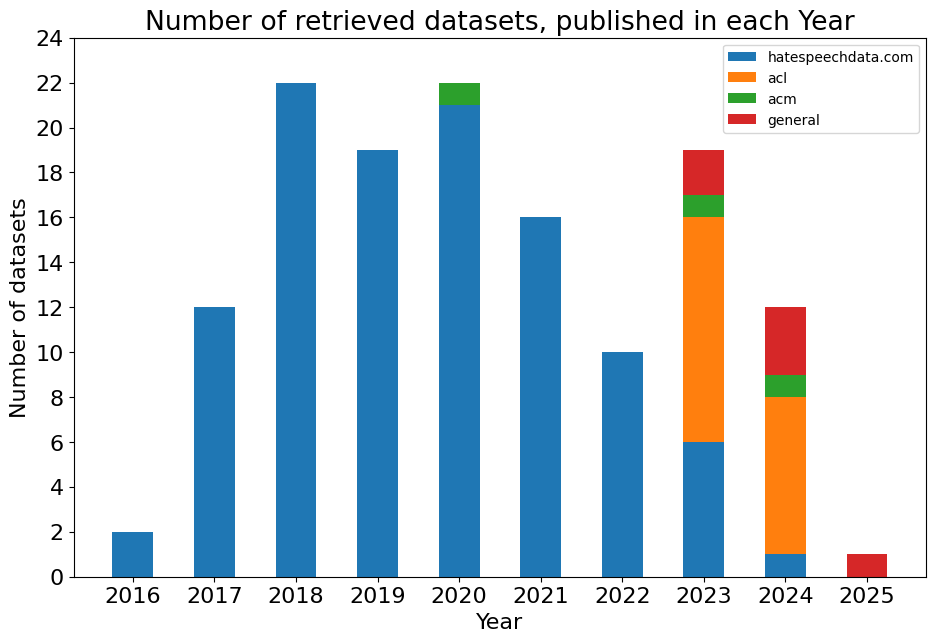}
    \caption{The number of datasets published in each year, split by source of retrieval.}
    \label{fig:frequency_count_by_years_by_sources}
\end{figure}

We treat substantially different datasets introduced within the same paper as distinct datasets \citep[e.g.][]{kumar-etal-2018-aggression}, as the datasets differ in both data sources and collection methods. In contrast, we regard \textsc{ETHOS} \cite{mollas_ethos_2022} as a single dataset despite its use of two data sources, since other aspects of its creation process remain consistent.
In total, we retrieved 135 distinct datasets across 36 languages. Figure~\ref{fig:frequency_count_by_years_by_sources} shows a breakdown of the number of datasets published in each year, split by source.

\section{Key Insights and Observations}
\label{sec_findings}

\subsection{Frames of Reference}
We begin by examining how the authors frame hate speech. Specifically, we look for explicit statements such as ``we define hate speech as...'' or ``hate speech is...''. Given the absence, and perhaps impossibility, of a universal definition \cite{vidgen-direction-2021,Poletto_Basile_Sanguinetti_Bosco_Patti_2021} and the heterogeneity of the designed tasks, we do not focus on measuring overlap or agreement between definitions. Instead, we identify key areas of coverage and commonly adopted definitions.


Of the 135 datasets, 23 (17\%) do not report a definition, and 71 (53\%) adopt prior definitions. The remaining 41 (30\%) state their own definitions. We analyse the definitions from three overlapping perspectives: 1) categorisation of hate speech into subtypes (e.g., racism, sexism, or categories such as threats and humiliation); 2) specification of the basis for hate (e.g., identities or group affiliations); and 3) referencing of intent (e.g., incite violence, harassment, or insult). Table \ref{tab:apx_def_break_down} presents the breakdown of datasets according to these aspects. Among the reported definitions, the basis for hate is most frequently highlighted (60\%), followed by subcategorisation (47\%) and intent (36\%).

\subsection{Goals}
We examine the designed goals of these datasets, i.e., the research objectives they were designed to achieve. Similar to our analysis of \textit{frames of reference}, we rely on signposting terms such as ``aim'', ``goal'', and ``to ...''. In a number of cases, we infer the aims based on contextual clues without the authors explicitly stating them.

We manually code the stated goals into eight categories: 1) promoting research, new directions, or underrepresented languages ($n=34$); 2) enabling comparison studies ($n=3$); 3) supporting automation or model development ($n=39$); 4) providing finer-grained annotations ($n=10$); 5) generating insights ($n=16$); 6) presenting new datasets and resources ($n=11$); 7) addressing research gaps and challenges ($n=28$); and 8) benchmarking ($n=20$). The goals and their associated datasets are listed in Table \ref{tab:apx_goal_breakdown}. This shows a considerable proportion of research focusses on automation and model development, exploring new directions in the field, and addressing known challenges.

\subsection{Languages}
Table \ref{tab:apx_lang_breakdown} shows the distribution of languages. By far, English has received the most attention. The next most frequently studied languages---Italian and German---lag behind by a sizeable margin. There are efforts focusing on multilingual capabilities, as indicated by the mixed-language datasets. Additionally, code-switching has gained traction as a research focus. However, even within code-switched datasets, English remains consistently present, receiving a large portion of attention.

Linguistic variations also play a role in dataset representation. Researchers distinguish between Brazilian Portuguese and European Portuguese, as well as between Mexican Spanish and European Spanish, to account for dialectal differences. Regional and creole languages \cite{Muysken_1995_study}, such as Singlish and Hinglish, are included but a strong English basis remains.

Contrary to \citet{tonneau-etal-2024-languages}, we did not observe a decline in English datasets' dominance. Instead, compared to non-English datasets, their proportion remains stable in years with more than three retrieved datasets. Possible reasons include different search scopes and methods.

\subsection{Data Collection}
Datasets are sourced using a variety of methods. 
Social media platforms dominate, with X/Twitter being the most prevalent data source ($n=70$). Other platforms include Facebook ($n=15$), YouTube 
($n=11$), and Reddit ($n=10$). Instagram ($n=2$) appears less frequently, likely due to its multimodality.
In contrast, traditional online forums are far less represented, with only a handful of datasets sourced from Gab ($n=4$) and Stormfront ($n=1$). News website comment sections also serve as a source of online hate ($n=13$). Additionally, three datasets originate from Wikipedia comments, and two from comments on online code repositories.
Beyond data collected ``from the wild'', some datasets are created ``in-house'' manually or synthetically ($n=10$). Other notable sources include language-specific platforms such as Sina Weibo \cite{JIANG2022swsr} and unconventional sources such as Russian subtitles from \textit{South Park} episodes \cite{saitov-derczynski-2021-abusive}. Table \ref{tab:apx_src_breakdown} lists these sources with their respective datasets.

The next step in the dataset creation pipeline is selecting datapoints for annotation. Researchers typically extract a subset of data from a larger corpus. Alternatively, a simpler one-step approach is employed, such as using keyword-based search to directly retrieve relevant instances.
We identify three primary techniques for data selection:
1) \textbf{Keyword-based sampling} ($n=73$): searching for relevant content using specific keywords and hashtags. It is the most common method.
2) \textbf{Keypage-based sampling} ($n=26$): focusses on specific recipients or platforms where hate speech is likely to occur. For instance, researchers collect data from key subreddits, Facebook pages, or Twitter accounts by selecting \textit{incoming} comments or tweets.
3) \textbf{Keyuser-based sampling} ($n=25$): unlike keypage-based selection, this technique focusses on the sender rather than the recipient. High-profile users are identified and their \textit{outgoing} comments or tweets are collected.

A subset of datasets ($n=7$) employ heuristic-based selection methods, applying thresholds to scores generated by external models. These models may be trained on a smaller dataset \cite{kennedy_2020_constructing} or leverage industry solutions such as PerspectiveAPI \citep[e.g.,][]{ElSherief_Nilizadeh_Nguyen_Vigna_Belding_2018, Sarker_2023_automated}.
\citet{kirk-etal-2023-semeval} introduce a unique approach using the score differential between two models as a selection criterion, making it the only dataset to employ a differential-based method.

All but one of the very large datasets ($n=7$), which contain entries numbering in the millions, do not not use any filtering. Instead, they are comments collected entirely from their respective hosting platforms with their moderation decision. The exception is from \citet{Borkan_2019_muanced}, which is a synthetic dataset. 

In terms of languages, geolocation filter ($n=5$) is commonly used to retrieve language-specific entries, besides data specific sources. 
Other filtering methods include random sampling \cite{Wulczyn_2017_ex, moon-etal-2020-beep, coltekin-2020-corpus, Kennedy_2022_gab}, filtering based on topic \cite{Pelle_2017_offensive, Madhu_2023_detecting}, and an active-learning-like method \cite{mollas_ethos_2022}.

We note many datasets ($n=47$) use multiple selection methods. When combined, these methods can function either as logical conjunction, i.e.\ datapoints must satisfy all the requirements to be included, or a logical disjunction, i.e.\ datapoints are selected if they satisfy at least one requirement. 

\subsection{Annotation}

\subsubsection{Task}
Hate speech detection can be formalised in various ways as a classification task. These formalisations vary in their granularity, determined by dataset curators' priorities and goals.
The simplest and most straightforward approach is binary classification ($n=34$), where datasets adopt a basic hateful/aggressive/toxic/abusive-or-not framework. While this is easy to implement and operationalise, it lacks nuance, failing to capture meaningful distinctions and subcategories within hate. 

Building on the binary classification framework, some datasets ($n=24$) adopt a multi-class classification approach, where each instance is assigned a single label from multiple ($>2$) mutually exclusive categories. This framework provides greater granularity, but it assumes clear-cut distinctions between categories, which may not always be compatible with the ambiguity introduced by edge cases and contexts. For instance, intersectional identities cannot be adequately expressed under this framework.
As a result, a model trained by these instances may be biased, as some identities are systematically underrepresented.

Further relaxing the assumption of rigid class boundaries, the multi-label framework ($n=4$) allows an instance to be assigned multiple applicable labels. In this approach, labels are organised in a flat structure, meaning they are mutually independent and not hierarchically related.

\begin{figure*}[ht] 
    \centering
        \begin{forest}
          for tree={
            grow=east,
            parent anchor=east,
            child anchor=west,
            edge={draw, thick},
            l sep=10mm
          }
          [Is it hateful?
            [No
              [Non-hate]
            ]
            [Yes
              [Targeted identities
                [African people]
                [Women]
                [LBGTQ+ communities]
                [Refugees]
                [Other]
              ]
              [Aggressiveness 
                [Strong]
                [Weak]
                [None]
              ]
              [Target type
                [Individual]
                [Group]
                [Other]
              ]
            ]
          ]
        \end{forest}
    \caption{A prototypical hierarchical categorisation of hate speech taxonomy.}
    \label{fig:hate_speech_prototypical}
\end{figure*}
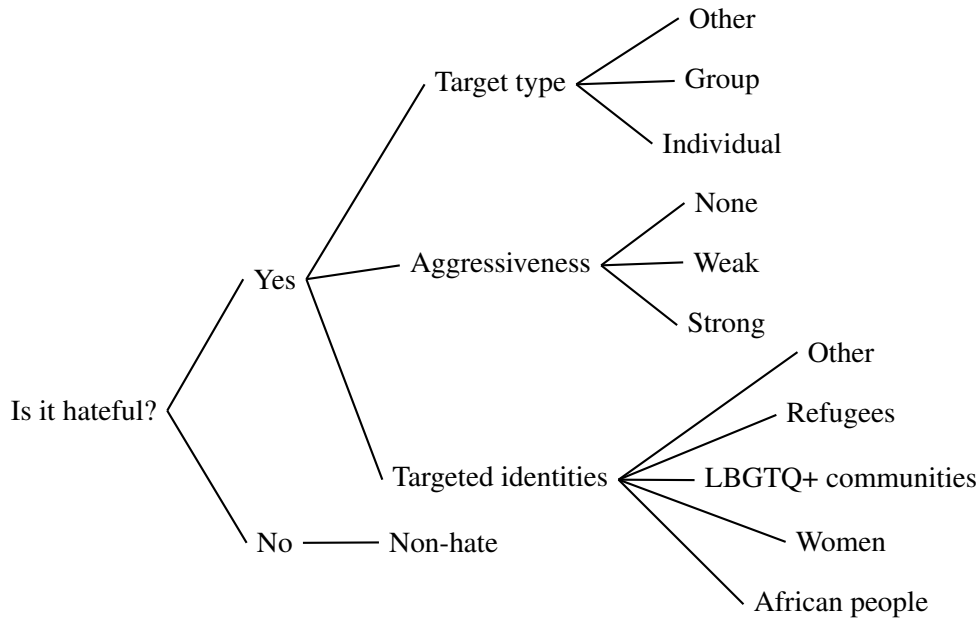

Labels can also be organised hierarchically ($n=54$), where labels are more structured, and can be tailored towards different levels of granularity. A well-defined taxonomy is essential to this framework. Notably, almost all ($n=43$) hierarchical datasets rely on an initial binary classification, where the root level question is a binary one. While this approach address the granularity problem, it also inherits the shortcomings of binary classification such as oversimplification. Figure \ref{fig:hate_speech_prototypical} depicts a prototypical hierarchical taxonomy.

We also identify another type of structure, which we refer to as a ``parallel'' structure ($n=7$). Unlike hierarchical frameworks that impose a single top-down taxonomy, parallel structures decouple multiple top-level concepts, allowing each to have its own independent internal structure. This approach provides greater flexibility in capturing different aspects of hate speech, as distinct dimensions can be subcategorised separately. For example, \citet{ousidhoum-etal-2019-multilingual} apply five classification taxonomies in parallel, covering directness, hostility type (including \textit{none}), target, group, and sentiment. 

Other types of formalisation include token-level classification \cite{pamungkas-etal-2020-really-fixed, pavlopoulos-etal-2021-semeval, saker2023toxispanseexplainabletoxicitydetection}. This approach offers more interpretability, but puts emphasis on inter-annotator agreement in relation to span boundaries. 

Each of these frameworks operationalises different ideal types, emphasising certain aspects of hate while overlooking others. No single framework fully captures the complexity of hate speech. 
Moreover, even when two datasets adopt the same framework, they may still show inconsistencies due to the differing underlying ideal types of hate, meaning that the apparent similarity in classification structure can be misleading, as differences in these ideal types are not immediately apparent.
Thus, a reflexive approach to dataset design, acknowledging and documenting these trade-offs, can lead to more effective and transparent datasets.

\subsubsection{Annotators}
The majority of the datasets use multiple annotators to label each example, while 13 have only one annotator attending to each example at some stage of annotation. However, in some cases multiple annotators are not feasible, for example when annotators are asked to \textit{construct} sentences \cite{goldzycher-etal-2024-improving}, rather than label them (Table \ref{tab:apx_anno_breakdown}).

Subsetting is a popular method to manage multiple annotators, where a (proper) subset of annotators from a pool is assigned to each example ($n=29$), while others ($n=47$) assign every annotator to every example. Crowdsourcing ($n=29$) is a special case of subsetting, where the annotator pool is large and not manually selected.

Among datasets with annotator subsetting, the pool sizes range from as few as three annotators \cite{pamungkas-etal-2020-really-fixed} to 50 \cite{Romim_2021_hate_bengali}. Most assign two annotators per instance, though some have up to five. For datasets without subsetting, the highest number of annotators assigned to an example is seven \cite{pavlopoulos-etal-2021-semeval}.

Smaller, hand-picked pools can increase annotation consistency, as researchers can enforce a uniform ideal type through additional training and moderation meetings, complementing prescriptive guidelines \cite{rottger-etal-2022-two}. In contrast, crowdsourcing makes large annotator pools more accessible, potentially increasing demographic diversity, but this is not always guaranteed \cite{tonneau-etal-2024-languages}. 
A larger pool is better suited for descriptive guidelines, which aim to capture the diversity of human opinions without imposing a predefined ideal type \cite{rottger-etal-2022-two}. However, under such settings, care must be taken to ensure actual diversity. Transparent reporting of annotator demographics is also vital in datasets with large annotator pools to assess potential biases and ensure a true representation of diverse ideal types.
\subsubsection{Annotator Demographics}
More than half of the datasets ($n=78$) do not report annotator demographics. Among those that do, the most commonly mentioned attributes are age ($n=33$), gender ($n=33$), and language ($n=32$). Other reported characteristics include education level ($n=18$) and location-based information such as nationality ($n=18$). A smaller number reference sexual orientation ($n=6$), proxies of socio-economic status (e.g., profession, income) ($n=10$), political leanings ($n=3$), or annotators’ prior experience with the subject matter, social media, or online abuse ($n=7$). Table \ref{tab:apx_anno_demo_breakdown} lists a subset of these dimensions. 

\subsubsection{Disagreements}
Most datasets aggregate multiple annotations into a single ground truth label. The utility of this step depends on the dataset's goal. For prescriptive guidelines, where a unified interpretation is intended, assigning a gold label is appropriate. However, for descriptive guidelines that aim to capture the diversity of human judgments, enforcing a single label is counterproductive \cite{rottger-etal-2022-two}.

To obtain gold labels, many datasets ($n=48$) use a simple majority rule, while some ($n=27$) involve additional annotators outside the original pool. Eight datasets resolve disagreements through moderation meetings. Other approaches include positive-class tie-breaking strategy \cite{gao-huang-2017-detecting}, and different positive threshold, where the positive label is assigned if positive annotations exceed a threshold \cite{leite-etal-2020-toxic, assenmacher2021textttrpmod} (Table \ref{tab:apx_aggre_breakdown}). Some datasets ($n=9$) discard instances with disagreement. However, this approach risks losing difficult and ambiguous cases, which can better capture real-world ambiguities, and may reinforce bias.

\subsubsection{Quality Assurance}
As a final dimension, we examine quality assurance (QA) measures, an often-overlooked aspect in previous surveys. We focus on the steps taken, if any, to ensure dataset quality. Around half of the datasets ($n=69$) do not report or are unclear about their QA procedures. Of those that do, we observe a relatively even distribution across approaches.

Before annotation, some crowdsourced datasets ($n=10$) select their annotators based on performance metrics (e.g.\ approval rate) as well as other data such as geo-location. 
Some incorporate on-boarding training ($n=13$), which may involve a trial where annotators label a small subset of the data \citep[e.g.][]{Golbeck_2017_large}. Twenty-four datasets employ moderation meetings, though only 10 explicitly mention refining guidelines based on discussions. 
Annotator tests are also employed by a number of datasets ($n=12$). These tests can be embedded in the annotation in the form of hidden tests and attention checks, or during onboarding, where annotators that fail a screening test are rejected\cite{assenmacher2021textttrpmod, lee-etal-2024-exploring-cross}.

Post-annotation QA includes external validation: ten datasets invite external experts to validate a subset of the annotation. 
Some datasets \citep[e.g.][]{pavlopoulos-etal-2017-deep, WiegandSiegelRuppenhofer2019} use external annotation and disagreement rates as a proxy for quality. This practice assumes a prescriptive guideline and goal, as high disagreement can still indicate high quality annotation under a descriptive framework \cite{lee-etal-2024-exploring-cross}.

\subsection{Ethics}

Of the papers reviewed, only 14 explicitly revealed they had approval or exemption from an Institutional Review Board (IRB) or ethics committee. A further 27 papers discussed ethical matters, such as anonymisation, but did not reveal if the research had undergone a review process. We note the exclusion of an ethics discussion does not mean the research was not reviewed, or imply that the research was not undertaken ethically. We notice a positive trend, with most of the more recent papers are least partly addressing ethical issues, indicating a growing recognition of the importance of ethics within the research community. 

By far the most discussed ethical concern was anonymisation ($n=21$). One of two approaches are commonly used for anonymisation when releasing datasets, as noted by~\citet{cercas-curry-etal-2021-convabuse}. The first approach is to only make an ID (e.g.\ Tweet ID) available, so that if a user or platform subsequently deletes a post it is no longer available. The second is to make the contents available, but to strip out any identifying information. The possibility that the datasets could be misused was considered in 11 papers, however it was noted that the benefits of the research typically outweighed any potential harm. Some researchers do not make their datasets available due to concerns about misuse~\cite{Golbeck_2017_large,Steffen_2023_codes,vargas-etal-2024-hausahate,Wijesiriwardene_2020_alone}, while others stipulate restrictions on use~\cite{assenmacher2021textttrpmod,fortuna-etal-2019-hierarchically,lee-etal-2024-exploring-cross}. 

The well-being of annotators, participants, and researchers was discussed in nine papers. Mitigations included allowing annotators to leave at any time~\cite{qian-etal-2019-benchmark,vasquez-etal-2023-homo}, making mental health support available~\cite{kirk-etal-2022-hatemoji,lee-etal-2024-exploring-cross,vidgen-etal-2021-introducing-fixed}, and briefing sessions and regular check-ins~\cite{kirk-etal-2022-hatemoji,kirk-etal-2023-semeval}. Eight papers also discussed the recruitment of annotators and participants, mainly in relation to compensation. To protect readers and to avoid the perpetuation of harms, authors refrained from providing direct quotes~\cite{cignarella-etal-2024-queereotypes,kirk-etal-2023-semeval,vidgen-direction-2021}, and provided content warnings~\cite{kirk-etal-2022-hatemoji,kirk-etal-2023-semeval}. 

Only one paper discussed environmental impacts, disclosing the energy sources for their computing clusters~\cite{castillo-lopez-etal-2023-analyzing}. In the future, we anticipate this will become a more prominent consideration, alongside more frequent use of LLMs and awareness of their environmental footprint.

\section{Discussions}
\label{sec_discussion}
\paragraph{A Reflexive Approach}
As hate speech detection inherently involves value judgements, it is crucial for researchers to adopt a reflexive approach throughout the dataset curation process, where the ideal types of hate and curatorial stances are critically examined and reported. In a prescriptive paradigm where disagreements and subjectivity are discouraged, the frame of references of the researchers can still shape their ideal-typical conceptualisation of the categories and definitions. Therefore, researchers must critically examine and document their own value judgements and frame of references as these ultimately shape the annotated datasets and trained models. By making these aspects explicit, researchers can promote transparency and allow for a more nuanced understanding of goal-driven ideal-typical constructs. 

\paragraph{Annotator Composition}
We note the interplay between annotator composition and the author's ideal-typical conceptualisations. Datasets with smaller, hand-picked annotator pools can more easily enforce a uniform ideal type through targeted training and discussions. This approach is more suited for prescriptive guidelines. Conversely, crowdsourced datasets can capture greater diversity, aligning better with descriptive guidelines. 
However, the persistent underreporting of crowdsourcing annotator demographics presents a challenge in assessing the diversity of captured opinions.

\paragraph{Annotation Aggregation}
While many datasets rely on majority voting, this method relies on two key assumptions: 1) ground truth is both obtainable and desirable, and 2) annotator consensus reflects this ground truth. Whether these assumptions hold depends on the operational framework. In a descriptive paradigm, aggregating annotations removes the diversity of responses rather than captures it. Additionally, majority voting leaves the underlying sources of disagreement unexamined, further introducing noise. Alternative approaches such as moderation meetings provide a more robust approach for resolving disagreements but are underutilised. Furthermore, datasets that discard instances with disagreement risk removing ambiguous cases, leading to an oversimplification of the task, which may reinforce existing biases.

\paragraph{Application of Ideal Types}
In this paper, we draw on Weber's notion of ideal types not as categories, but as interpretive lenses reflecting the dataset creators' conceptualisations. In principle, there could be as many ideal types as there are datasets, with each remaining valid within its own context.  Rather than attempting to force consensus, the notion of ideal types foregrounds and emphasise the importance of this diversity in curatorial stances. 

Furthermore, we suggest the use of Weber's ideal types of social action to interpret hate speech content. While they have not been used as categories to which each dataset is assigned, they can be used as analytical heuristics to interpret the socio-political underpinnings and motivations embedded in these datasets.  
For instance, \textsc{PubFigs-L}~\cite{Yuan_2025_generalizing} is a set of manually labelled tweets from 15 American political public figures across the political spectrum. 
The authors uncover six main themes in hateful and abusive speech: Islam, women, race and ethnicity, immigration and refugees, terrorism and extremism, and American politics \cite{Yuan_2025_generalizing}. Through a Weberian lens, such speech can be goal-rational, strategically used to further political agendas, or value-rational, such as religiously motivated hate. Affectual speech aligns with the dataset’s category of abuse, distinguishing identity-based hate from emotionally driven personal attacks. The authors also implicitly acknowledge traditional hate speech by noting the presence of covert and implicit hate.

Interpreting using ideal types allows researchers to better understand the heterogeneous curatorial decisions, and better account for the plural underpinnings that motivate hate speech content.

\section{Conclusion}
Through a Weberian lens, we examine hate speech datasets through Max Weber's ideal types of social action to understand the socio-political underpinnings. We illustrate examples of goal-rational hate, where political figures use hate and abusive language to mobilise the public for political gain, and value-rational hate, where hate speech is driven by ideological beliefs. Moreover, affectual hate can be attacks driven by emotions such as frustration and anger, while traditional hate speech is often normalised and implicit. 
These ideal types offer a theoretical grounding to the operationalisation of hate speech while acknowledging the diversity of design choices of researchers.
Our analysis highlights how dataset construction is shaped by various factors, including the researchers' frame of reference and goal, which in turn influence key design decisions. We advocate for a reflexive approach to dataset construction in which researchers critically examine their own assumptions, operationalisation choices, and the socio-political contexts that shape their work.
\section*{Limitations}
Our study primarily focusses on publicly available datasets, which may not fully represent the diversity of methodologies used in industry or private research. Second, while we examine key aspects such as frames of reference, goals, languages etc., we do not perform empirical evaluations of annotation quality or dataset performance in downstream tasks. Additionally, our discussion of ideal types and annotation paradigms is necessarily interpretative, and alternative theoretical frameworks could yield different viewpoints. 
\section*{Acknowledgments}

This paper is part of a project that has received funding from the European Research Council (ERC) under the European Union’s Horizon 2020 research and innovation programme (grant agreement No 949127).

\bibliography{anthology, custom}

\appendix

\section{Appendix}
\label{sec:appendix}

\clearpage
\onecolumn
\subsection{Breakdowns of Reviewed Datasets}

\renewcommand{\tabularxcolumn}[1]{m{#1}}
\begin{table}[ht]
\centering
\begin{tabularx}{\textwidth}{@{}Xcccc@{}}
\toprule
\textbf{Datasets} & \textbf{Subcategories} & \textbf{Basis} & \textbf{Intent} & \textbf{Total} \\ 
\midrule
\small\citet{jha-mamidi-2017-compliment-fixed, Salminen_Almerekhi_Milenković_Jung_An_Kwak_Jansen_2018, WiegandSiegelRuppenhofer2019, sprugnoli-etal-2018-creating, ousidhoum-etal-2019-multilingual, Borkan_2019_muanced, Shekhar_Pranjić_Pollak_Pelicon_Purver_2020, sigurbergsson-derczynski-2020-offensive, caselli-etal-2020-feel-fixed, pavlopoulos-etal-2020-toxicity, albanyan-etal-2023-counterhate, korre-etal-2023-harmful, seo-etal-2024-kocommongen, ng-etal-2024-sghatecheck} &
  \fullmoon & \fullmoon & \fullmoon & 17 \\ 

\spacerow

\small\citet{Golbeck_2017_large, ljubesic-etal-2018-datasets, zampieri-etal-2019-predicting, Shekhar_Pranjić_Pollak_Pelicon_Purver_2020, pitenis-etal-2020-offensive, leite-etal-2020-toxic, saitov-derczynski-2021-abusive, nurce2022detectingabusivealbanian, shekhar-etal-2022-coral-fixed, saker2023toxispanseexplainabletoxicitydetection, Sarker_2023_automated, raihan-etal-2023-offensive} & \newmoon  & \fullmoon & \fullmoon & 13 \\

\spacerow

\small\citet{ross_2016_measuring, Pelle_2017_offensive, Fersini_2018_overview, ElSherief_Nilizadeh_Nguyen_Vigna_Belding_2018, chung-etal-2019-conan, qian-etal-2019-benchmark, basile-etal-2019-semeval, ibrohim-budi-2019-multi, kennedy_2020_constructing, coltekin-2020-corpus, vidgen-etal-2021-learning, grimminger-klinger-2021-hate, rottger-etal-2021-hatecheck, mollas_ethos_2022, ollagnier-etal-2022-cyberagressionado, Trajano_Bordini_Vieira_2024, kirk-etal-2023-semeval, Steffen_2023_codes, goldzycher-etal-2024-improving} & \fullmoon & \newmoon  & \fullmoon & 27 \\

\spacerow

\small\citet{Bretschneider2017DetectingOS, alvarez2018overview, suryawanshi-etal-2020-multimodal, Wijesiriwardene_2020_alone, kurrek-etal-2020-towards, caselli-etal-2021-dalc-fixed, Kennedy_2022_gab, park-etal-2023-uncovering, rawat-etal-2023-modelling}& \fullmoon & \fullmoon & \newmoon  & 11 \\

\spacerow

\small\citet{Rezvan_2018_quality, Samory_Sen_Kohne_Flöck_Wagner_2021}& \newmoon  & \newmoon  & \fullmoon & 2 \\

\spacerow

\small\citet{waseem-hovy-2016-hateful-fixed, waseem-2016-racist-fixed, mubarak-etal-2017-abusive}& \newmoon  & \fullmoon & \newmoon  & 3 \\

\spacerow

\small\citet{gao-huang-2017-detecting, Alfina_2017_indonesian, de-gibert-etal-2018-hate, mathur-etal-2018-offend-fixed, Ptaszynski_Pieciukiewicz_Dybala_2019, fortuna-etal-2019-hierarchically, Gomez_2020_exploring, Romim_2021_hate_bengali, toraman-etal-2022-large, kirk-etal-2022-hatemoji, demus-etal-2022-comprehensive-fixed, castillo-lopez-etal-2023-analyzing, Saeed_2023_detection, Das_Raj_Saha_Mathew_Gupta_Mukherjee_2023}& \fullmoon & \newmoon  & \newmoon  & 16 \\

\spacerow

\small\citet{Albadi_2018_are, Founta_Djouvas_Chatzakou_Leontiadis_Blackburn_Stringhini_Vakali_Sirivianos_Kourtellis_2018, sanguinetti-etal-2018-italian, bosco2018overview, mulki-etal-2019-l, Mandl_2019_hasoc, pamungkas-etal-2020-really-fixed, vidgen-etal-2020-detecting, rizwan-etal-2020-hate, bhardwaj2020hostilitydetectiondatasethindi, moon-etal-2020-beep, Fersini_Nozza_Rosso_2020, mulki-ghanem-2021-mi, vidgen-etal-2021-introducing-fixed, assenmacher2021textttrpmod, JIANG2022swsr, ilevbare-etal-2024-ekohate, singh_2024_mimic, Yuan_2025_generalizing} & \newmoon  & \newmoon  & \newmoon  & 22 \\

\spacerow
         
\small\citet{mubarak-etal-2017-abusive, Wulczyn_2017_ex, pavlopoulos-etal-2017-deep, Alakrot_2018_dataset, kumar-etal-2018-aggression, bohra-etal-2018-dataset, Ibrohim_2018_preliminaries, zueva-etal-2020-reducing, raman_2020_stress, zeinert-etal-2021-annotating, cercas-curry-etal-2021-convabuse, pavlopoulos-etal-2021-semeval, fanton-etal-2021-human-fixed, Mathew_Saha_Yimam_Biemann_Goyal_Mukherjee_2021, vasquez-etal-2023-homo, Madhu_2023_detecting, cignarella-etal-2024-queereotypes, vargas-etal-2024-hausahate, dementieva-etal-2024-toxicity, Ferreira_2024_towards, lee-etal-2024-exploring-cross, Sreelakshmi_2024_detection}& \multicolumn{3}{c}{\textit{not reported}} & 24 \\ 
\bottomrule
\end{tabularx}
\caption{How the definitions are constructed in each dataset. \fullmoon: not present, \newmoon: present. Note that one paper may introduce multiple datasets. The number of references and the number of datasets are not necessarily equal.}
\label{tab:apx_def_break_down}
\end{table}

\begin{table}[ht]
\begin{tabularx}{\textwidth}{@{}>{\centering\arraybackslash}m{4cm}Xc@{}}
\toprule
\textbf{Coded goals}    & \multicolumn{1}{c}{\textbf{Datasets}} & \multicolumn{1}{c}{\textbf{Count}} \\
\midrule
Promoting research, new directions, or underrepresented languages & \small\citet{waseem-hovy-2016-hateful-fixed, Pelle_2017_offensive, WiegandSiegelRuppenhofer2019, kumar-etal-2018-aggression, bohra-etal-2018-dataset, bosco2018overview, alvarez2018overview, Mandl_2019_hasoc, Ptaszynski_Pieciukiewicz_Dybala_2019, fortuna-etal-2019-hierarchically, kennedy_2020_constructing, Gomez_2020_exploring, moon-etal-2020-beep, Fersini_Nozza_Rosso_2020, leite-etal-2020-toxic, coltekin-2020-corpus, rizwan-etal-2020-hate, raman_2020_stress, saitov-derczynski-2021-abusive, Trajano_Bordini_Vieira_2024, rawat-etal-2023-modelling, vasquez-etal-2023-homo, Steffen_2023_codes, raihan-etal-2023-offensive, Saeed_2023_detection, ilevbare-etal-2024-ekohate, vargas-etal-2024-hausahate, dementieva-etal-2024-toxicity}  &  34  \\ \spacerow
Enabling comparison studies             & \small\citet{waseem-2016-racist-fixed, basile-etal-2019-semeval} & 3 \\ \spacerow
Supporting automation or model development & \small\citet{mubarak-etal-2017-abusive, Golbeck_2017_large, Wulczyn_2017_ex, pavlopoulos-etal-2017-deep, Alfina_2017_indonesian, Pelle_2017_offensive, Alakrot_2018_dataset, sanguinetti-etal-2018-italian, qian-etal-2019-benchmark, Shekhar_Pranjić_Pollak_Pelicon_Purver_2020, sigurbergsson-derczynski-2020-offensive, vidgen-etal-2020-detecting, pavlopoulos-etal-2020-toxicity, zeinert-etal-2021-annotating, Samory_Sen_Kohne_Flöck_Wagner_2021, pavlopoulos-etal-2021-semeval, vidgen-etal-2021-introducing-fixed, mollas_ethos_2022, nurce2022detectingabusivealbanian, kirk-etal-2022-hatemoji, saker2023toxispanseexplainabletoxicitydetection, Sarker_2023_automated, Trajano_Bordini_Vieira_2024, park-etal-2023-uncovering, kirk-etal-2023-semeval, Saeed_2023_detection, Das_Raj_Saha_Mathew_Gupta_Mukherjee_2023, cignarella-etal-2024-queereotypes, Yuan_2025_generalizing} & 39 \\ \spacerow
Providing finer-grained annotations       & \small\citet{Davidson_Warmsley_Macy_Weber_2017, Fersini_2018_overview, Founta_Djouvas_Chatzakou_Leontiadis_Blackburn_Stringhini_Vakali_Sirivianos_Kourtellis_2018, zampieri-etal-2019-predicting, vidgen-etal-2021-introducing-fixed, assenmacher2021textttrpmod, shekhar-etal-2022-coral-fixed, Kennedy_2022_gab, demus-etal-2022-comprehensive-fixed} & 10 \\ \spacerow
Generating insights                       & \small\citet{Golbeck_2017_large, ross_2016_measuring, ElSherief_Nilizadeh_Nguyen_Vigna_Belding_2018, Salminen_Almerekhi_Milenković_Jung_An_Kwak_Jansen_2018, sprugnoli-etal-2018-creating, Ptaszynski_Pieciukiewicz_Dybala_2019, pamungkas-etal-2020-really-fixed, pavlopoulos-etal-2020-toxicity, cercas-curry-etal-2021-convabuse, grimminger-klinger-2021-hate, assenmacher2021textttrpmod, JIANG2022swsr, albanyan-etal-2023-counterhate, Madhu_2023_detecting, cignarella-etal-2024-queereotypes} & 16 \\ \spacerow
Presenting new datasets and resources      & \small\citet{Rezvan_2018_quality, chung-etal-2019-conan, pitenis-etal-2020-offensive, bhardwaj2020hostilitydetectiondatasethindi, moon-etal-2020-beep, Romim_2021_hate_bengali, caselli-etal-2021-dalc-fixed, grimminger-klinger-2021-hate, fanton-etal-2021-human-fixed} & 11 \\ \spacerow
Addressing research gaps and challenges    & \small\citet{gao-huang-2017-detecting, jha-mamidi-2017-compliment-fixed, Albadi_2018_are, ljubesic-etal-2018-datasets, de-gibert-etal-2018-hate, mathur-etal-2018-offend-fixed, Ibrohim_2018_preliminaries, Borkan_2019_muanced, ibrohim-budi-2019-multi, caselli-etal-2020-feel-fixed, suryawanshi-etal-2020-multimodal, Fersini_Nozza_Rosso_2020, zueva-etal-2020-reducing, vidgen-etal-2021-learning, fanton-etal-2021-human-fixed, Kennedy_2022_gab, ollagnier-etal-2022-cyberagressionado, kirk-etal-2023-semeval, Das_Raj_Saha_Mathew_Gupta_Mukherjee_2023, Madhu_2023_detecting, goldzycher-etal-2024-improving, ng-etal-2024-sghatecheck, singh_2024_mimic, Ferreira_2024_towards, lee-etal-2024-exploring-cross, Yuan_2025_generalizing} & 28 \\ \spacerow
Benchmarking                    & \small\citet{Bretschneider2017DetectingOS, sanguinetti-etal-2018-italian, ousidhoum-etal-2019-multilingual, mulki-etal-2019-l, kurrek-etal-2020-towards, moon-etal-2020-beep, mulki-ghanem-2021-mi, rottger-etal-2021-hatecheck, Mathew_Saha_Yimam_Biemann_Goyal_Mukherjee_2021, shekhar-etal-2022-coral-fixed, toraman-etal-2022-large, kirk-etal-2022-hatemoji, korre-etal-2023-harmful, castillo-lopez-etal-2023-analyzing, seo-etal-2024-kocommongen, Sreelakshmi_2024_detection} & 20 \\ \spacerow
\textit{not reported}           & \small\citet{Wijesiriwardene_2020_alone} & 1 \\
\bottomrule
\end{tabularx}%
\caption{Breakdown of datasets by goal.}
\label{tab:apx_goal_breakdown}
\end{table}

\begin{table}[ht]
\begin{tabularx}{\textwidth}{@{}>{\centering\arraybackslash}m{3.5cm}Xc@{}}
\toprule
\textbf{Languages}    & \multicolumn{1}{c}{\textbf{Datasets}} & \multicolumn{1}{c}{\textbf{Count}} \\
\midrule
English & \small\citet{waseem-hovy-2016-hateful-fixed, waseem-2016-racist-fixed, Davidson_Warmsley_Macy_Weber_2017, gao-huang-2017-detecting, jha-mamidi-2017-compliment-fixed, Golbeck_2017_large, Wulczyn_2017_ex, de-gibert-etal-2018-hate, Fersini_2018_overview, ElSherief_Nilizadeh_Nguyen_Vigna_Belding_2018, Founta_Djouvas_Chatzakou_Leontiadis_Blackburn_Stringhini_Vakali_Sirivianos_Kourtellis_2018, Rezvan_2018_quality, Salminen_Almerekhi_Milenković_Jung_An_Kwak_Jansen_2018, ousidhoum-etal-2019-multilingual, zampieri-etal-2019-predicting, Borkan_2019_muanced, chung-etal-2019-conan, qian-etal-2019-benchmark, basile-etal-2019-semeval, Mandl_2019_hasoc, kennedy_2020_constructing, caselli-etal-2020-feel-fixed, pamungkas-etal-2020-really-fixed, suryawanshi-etal-2020-multimodal, Wijesiriwardene_2020_alone, kurrek-etal-2020-towards, Gomez_2020_exploring, vidgen-etal-2020-detecting, pavlopoulos-etal-2020-toxicity, raman_2020_stress, cercas-curry-etal-2021-convabuse, vidgen-etal-2021-learning, Samory_Sen_Kohne_Flöck_Wagner_2021, grimminger-klinger-2021-hate, rottger-etal-2021-hatecheck, pavlopoulos-etal-2021-semeval, fanton-etal-2021-human-fixed, Mathew_Saha_Yimam_Biemann_Goyal_Mukherjee_2021, vidgen-etal-2021-introducing-fixed, mollas_ethos_2022, toraman-etal-2022-large, kirk-etal-2022-hatemoji, Kennedy_2022_gab, albanyan-etal-2023-counterhate, saker2023toxispanseexplainabletoxicitydetection, Sarker_2023_automated, korre-etal-2023-harmful, park-etal-2023-uncovering, kirk-etal-2023-semeval, Das_Raj_Saha_Mathew_Gupta_Mukherjee_2023, lee-etal-2024-exploring-cross, Yuan_2025_generalizing}  &  55  \\ \spacerow
Italian  & \small\citet{sanguinetti-etal-2018-italian, bosco2018overview, sprugnoli-etal-2018-creating, chung-etal-2019-conan, Fersini_Nozza_Rosso_2020, cignarella-etal-2024-queereotypes} & 8 \\ \spacerow

German  & \small\citet{ross_2016_measuring, Bretschneider2017DetectingOS, WiegandSiegelRuppenhofer2019, Mandl_2019_hasoc, assenmacher2021textttrpmod, demus-etal-2022-comprehensive-fixed, Steffen_2023_codes, goldzycher-etal-2024-improving} & 8 \\ \spacerow

Arabic & \small\citet{mubarak-etal-2017-abusive, Albadi_2018_are, Alakrot_2018_dataset, ousidhoum-etal-2019-multilingual} & 5 \\ \spacerow

Barzilian Portuguese       & \small\citet{Pelle_2017_offensive, leite-etal-2020-toxic, Trajano_Bordini_Vieira_2024} & 5 \\ \spacerow

Croatian       & \small\citet{ljubesic-etal-2018-datasets, Shekhar_Pranjić_Pollak_Pelicon_Purver_2020, shekhar-etal-2022-coral-fixed} & 4 \\ \spacerow

Spanish, French, Indonesian, Korean (3 each)       & \small\citet{Alfina_2017_indonesian, Fersini_2018_overview, Ibrohim_2018_preliminaries, ousidhoum-etal-2019-multilingual, chung-etal-2019-conan, basile-etal-2019-semeval, ibrohim-budi-2019-multi, moon-etal-2020-beep, ollagnier-etal-2022-cyberagressionado, park-etal-2023-uncovering, castillo-lopez-etal-2023-analyzing, seo-etal-2024-kocommongen} & 3 $\times$ 4 \\ \spacerow

Hindi, Danish, Turkish, Greek, Russian, Mexican Spanish, Portuguese (2 each)      & \small\citet{pavlopoulos-etal-2017-deep, alvarez2018overview, Mandl_2019_hasoc, fortuna-etal-2019-hierarchically, sigurbergsson-derczynski-2020-offensive, pitenis-etal-2020-offensive, bhardwaj2020hostilitydetectiondatasethindi, zueva-etal-2020-reducing, coltekin-2020-corpus, zeinert-etal-2021-annotating, saitov-derczynski-2021-abusive, toraman-etal-2022-large, vasquez-etal-2023-homo, Ferreira_2024_towards} & 2 $\times$ 7 \\ \spacerow

Slovenian, Levantine, Bengali, Dutch, Albanian, Chinese, Hinglish, Polish, Roman Urdu, Hausa, Ukrainian, Urdu (1 each)       & \small\citet{ljubesic-etal-2018-datasets, mathur-etal-2018-offend-fixed, mulki-etal-2019-l, Ptaszynski_Pieciukiewicz_Dybala_2019, rizwan-etal-2020-hate, Romim_2021_hate_bengali, caselli-etal-2021-dalc-fixed, nurce2022detectingabusivealbanian, JIANG2022swsr, Saeed_2023_detection, vargas-etal-2024-hausahate, dementieva-etal-2024-toxicity} & 1 $\times$ 12 \\ \spacerow

Mixed languages     & \small Estonian, Russian: \citet{Shekhar_Pranjić_Pollak_Pelicon_Purver_2020}; Arabic, Levantine: \citet{mulki-ghanem-2021-mi}; Singlish, Malay, and Tamil: \citet{ng-etal-2024-sghatecheck} & 3 \\ \spacerow

Code-switched languages & \small Hindi, English ($n=6$): \citet{kumar-etal-2018-aggression, bohra-etal-2018-dataset, rawat-etal-2023-modelling, Madhu_2023_detecting, singh_2024_mimic}; Malayalam, English ($n=1$): \citet{Sreelakshmi_2024_detection}; Bengali, English ($n=1$): \citet{raihan-etal-2023-offensive}; Yoruba, Naija, English ($n=1$): \citet{ilevbare-etal-2024-ekohate} & 9 \\ \spacerow
\bottomrule
\end{tabularx}%
\caption{Breakdown of datasets by language. Datasets labelled as ``mixed languages'' contain texts from multiple languages, but individual texts are not code-mixed. In contrast, ``code-switched datasets'' refer to datasets where individual entries exhibit code-switching.}
\label{tab:apx_lang_breakdown}
\end{table}

\begin{table}[ht]
\begin{tabularx}{\textwidth}{@{}>{\centering\arraybackslash}m{3.5cm}Xc@{}}
\toprule
\textbf{Source}    & \multicolumn{1}{c}{\textbf{Datasets}} & \multicolumn{1}{c}{\textbf{Count}} \\
\midrule
Twitter & \small\citet{waseem-hovy-2016-hateful-fixed, waseem-2016-racist-fixed, mubarak-etal-2017-abusive, Davidson_Warmsley_Macy_Weber_2017, jha-mamidi-2017-compliment-fixed, Golbeck_2017_large, ross_2016_measuring, Alfina_2017_indonesian, Albadi_2018_are, Fersini_2018_overview, ElSherief_Nilizadeh_Nguyen_Vigna_Belding_2018, Founta_Djouvas_Chatzakou_Leontiadis_Blackburn_Stringhini_Vakali_Sirivianos_Kourtellis_2018, Rezvan_2018_quality, WiegandSiegelRuppenhofer2019, kumar-etal-2018-aggression, mathur-etal-2018-offend-fixed, bohra-etal-2018-dataset, Ibrohim_2018_preliminaries, sanguinetti-etal-2018-italian, bosco2018overview, alvarez2018overview, ousidhoum-etal-2019-multilingual, mulki-etal-2019-l, zampieri-etal-2019-predicting, chung-etal-2019-conan, basile-etal-2019-semeval, Mandl_2019_hasoc, ibrohim-budi-2019-multi, Ptaszynski_Pieciukiewicz_Dybala_2019, fortuna-etal-2019-hierarchically, sigurbergsson-derczynski-2020-offensive, kennedy_2020_constructing, Wijesiriwardene_2020_alone, Gomez_2020_exploring, vidgen-etal-2020-detecting, pitenis-etal-2020-offensive, bhardwaj2020hostilitydetectiondatasethindi, Fersini_Nozza_Rosso_2020, leite-etal-2020-toxic, coltekin-2020-corpus, rizwan-etal-2020-hate, mulki-ghanem-2021-mi, zeinert-etal-2021-annotating, caselli-etal-2021-dalc-fixed, Samory_Sen_Kohne_Flöck_Wagner_2021, grimminger-klinger-2021-hate, Mathew_Saha_Yimam_Biemann_Goyal_Mukherjee_2021, toraman-etal-2022-large, kirk-etal-2022-hatemoji, demus-etal-2022-comprehensive-fixed, albanyan-etal-2023-counterhate, Trajano_Bordini_Vieira_2024, castillo-lopez-etal-2023-analyzing, rawat-etal-2023-modelling, vasquez-etal-2023-homo, Saeed_2023_detection, Madhu_2023_detecting, cignarella-etal-2024-queereotypes, ilevbare-etal-2024-ekohate, Ferreira_2024_towards, Yuan_2025_generalizing}  &  70  \\ \spacerow
Facebook  & \small\citet{Bretschneider2017DetectingOS, Salminen_Almerekhi_Milenković_Jung_An_Kwak_Jansen_2018, kumar-etal-2018-aggression, bosco2018overview, Mandl_2019_hasoc, sigurbergsson-derczynski-2020-offensive, bhardwaj2020hostilitydetectiondatasethindi, Romim_2021_hate_bengali, zeinert-etal-2021-annotating, raihan-etal-2023-offensive, cignarella-etal-2024-queereotypes, vargas-etal-2024-hausahate, singh_2024_mimic} & 15 \\ \spacerow

YouTube  & \small\citet{Alakrot_2018_dataset, Salminen_Almerekhi_Milenković_Jung_An_Kwak_Jansen_2018, kennedy_2020_constructing, Romim_2021_hate_bengali, mollas_ethos_2022, nurce2022detectingabusivealbanian, Trajano_Bordini_Vieira_2024, park-etal-2023-uncovering, lee-etal-2024-exploring-cross, Sreelakshmi_2024_detection} & 11 \\ \spacerow

Reddit & \small\citet{qian-etal-2019-benchmark, sigurbergsson-derczynski-2020-offensive, kennedy_2020_constructing, kurrek-etal-2020-towards, zeinert-etal-2021-annotating, vidgen-etal-2021-introducing-fixed, mollas_ethos_2022, kirk-etal-2023-semeval, singh_2024_mimic, lee-etal-2024-exploring-cross} & 10 \\ \spacerow

Instagram     & \small\citet{nurce2022detectingabusivealbanian, singh_2024_mimic} & 2 \\ \spacerow

Gab \& Stormfront       & \small\citet{de-gibert-etal-2018-hate, qian-etal-2019-benchmark, Mathew_Saha_Yimam_Biemann_Goyal_Mukherjee_2021, Kennedy_2022_gab, kirk-etal-2023-semeval} & 5 \\ \spacerow

Human Creation       & \small\citet{chung-etal-2019-conan, cercas-curry-etal-2021-convabuse, fanton-etal-2021-human-fixed, ollagnier-etal-2022-cyberagressionado, goldzycher-etal-2024-improving} & 7 \\ \spacerow

Synthetic    & \small\citet{vidgen-etal-2021-learning, rottger-etal-2021-hatecheck, kirk-etal-2022-hatemoji} & 3 \\ \spacerow

Existing datasets & \small\citet{caselli-etal-2020-feel-fixed, pamungkas-etal-2020-really-fixed, pavlopoulos-etal-2021-semeval, saker2023toxispanseexplainabletoxicitydetection, Trajano_Bordini_Vieira_2024, korre-etal-2023-harmful, seo-etal-2024-kocommongen, ng-etal-2024-sghatecheck, dementieva-etal-2024-toxicity, lee-etal-2024-exploring-cross} & 10 \\ \spacerow

Other    & \small  \citet{mubarak-etal-2017-abusive, gao-huang-2017-detecting, Wulczyn_2017_ex, pavlopoulos-etal-2017-deep, Pelle_2017_offensive, ljubesic-etal-2018-datasets, sprugnoli-etal-2018-creating, Borkan_2019_muanced, Shekhar_Pranjić_Pollak_Pelicon_Purver_2020, suryawanshi-etal-2020-multimodal, pavlopoulos-etal-2020-toxicity, moon-etal-2020-beep, zueva-etal-2020-reducing, raman_2020_stress, assenmacher2021textttrpmod, saitov-derczynski-2021-abusive, JIANG2022swsr, shekhar-etal-2022-coral-fixed, Sarker_2023_automated, Steffen_2023_codes, Das_Raj_Saha_Mathew_Gupta_Mukherjee_2023} & 24 \\ \spacerow

\bottomrule
\end{tabularx}%
\caption{Breakdown of datasets by data source.}
\label{tab:apx_src_breakdown}
\end{table}

\begin{table}[ht]
\begin{tabularx}{\textwidth}{@{}>{\centering\arraybackslash}m{3.5cm}Xc@{}}
\toprule
\textbf{Collection method}    & \multicolumn{1}{c}{\textbf{Datasets}} & \multicolumn{1}{c}{\textbf{Count}} \\
\midrule
Keyword-based & \small\citet{waseem-hovy-2016-hateful-fixed, mubarak-etal-2017-abusive, Davidson_Warmsley_Macy_Weber_2017, jha-mamidi-2017-compliment-fixed, Golbeck_2017_large, ross_2016_measuring, Alfina_2017_indonesian, Albadi_2018_are, Fersini_2018_overview, ElSherief_Nilizadeh_Nguyen_Vigna_Belding_2018, Rezvan_2018_quality, Salminen_Almerekhi_Milenković_Jung_An_Kwak_Jansen_2018, WiegandSiegelRuppenhofer2019, kumar-etal-2018-aggression, mathur-etal-2018-offend-fixed, bohra-etal-2018-dataset, Ibrohim_2018_preliminaries, sanguinetti-etal-2018-italian, bosco2018overview, alvarez2018overview, ousidhoum-etal-2019-multilingual, mulki-etal-2019-l, zampieri-etal-2019-predicting, qian-etal-2019-benchmark, basile-etal-2019-semeval, Mandl_2019_hasoc, ibrohim-budi-2019-multi, fortuna-etal-2019-hierarchically, sigurbergsson-derczynski-2020-offensive, pamungkas-etal-2020-really-fixed, Wijesiriwardene_2020_alone, kurrek-etal-2020-towards, Gomez_2020_exploring, vidgen-etal-2020-detecting, pitenis-etal-2020-offensive, bhardwaj2020hostilitydetectiondatasethindi, leite-etal-2020-toxic, rizwan-etal-2020-hate, Romim_2021_hate_bengali, zeinert-etal-2021-annotating, caselli-etal-2021-dalc-fixed, cercas-curry-etal-2021-convabuse, Samory_Sen_Kohne_Flöck_Wagner_2021, grimminger-klinger-2021-hate, Mathew_Saha_Yimam_Biemann_Goyal_Mukherjee_2021, JIANG2022swsr, toraman-etal-2022-large, kirk-etal-2022-hatemoji, demus-etal-2022-comprehensive-fixed, Trajano_Bordini_Vieira_2024, castillo-lopez-etal-2023-analyzing, rawat-etal-2023-modelling, kirk-etal-2023-semeval, vasquez-etal-2023-homo, raihan-etal-2023-offensive, Saeed_2023_detection, Das_Raj_Saha_Mathew_Gupta_Mukherjee_2023, cignarella-etal-2024-queereotypes, seo-etal-2024-kocommongen, vargas-etal-2024-hausahate, singh_2024_mimic, Ferreira_2024_towards, lee-etal-2024-exploring-cross}  &  73  \\ \spacerow
Keypage-based  & \small\citet{gao-huang-2017-detecting, Bretschneider2017DetectingOS, Alakrot_2018_dataset, Fersini_2018_overview, kumar-etal-2018-aggression, bosco2018overview, qian-etal-2019-benchmark, sigurbergsson-derczynski-2020-offensive, kennedy_2020_constructing, kurrek-etal-2020-towards, raman_2020_stress, mulki-ghanem-2021-mi, Romim_2021_hate_bengali, vidgen-etal-2021-introducing-fixed, nurce2022detectingabusivealbanian, Trajano_Bordini_Vieira_2024, park-etal-2023-uncovering, kirk-etal-2023-semeval, Steffen_2023_codes, raihan-etal-2023-offensive, cignarella-etal-2024-queereotypes, ilevbare-etal-2024-ekohate, vargas-etal-2024-hausahate, singh_2024_mimic} & 26 \\ \spacerow

Keyuser-based  & \small\citet{waseem-hovy-2016-hateful-fixed, Wulczyn_2017_ex, Fersini_2018_overview, ElSherief_Nilizadeh_Nguyen_Vigna_Belding_2018, WiegandSiegelRuppenhofer2019, mulki-etal-2019-l, basile-etal-2019-semeval, Mandl_2019_hasoc, Ptaszynski_Pieciukiewicz_Dybala_2019, fortuna-etal-2019-hierarchically, Wijesiriwardene_2020_alone, kurrek-etal-2020-towards, leite-etal-2020-toxic, rizwan-etal-2020-hate, zeinert-etal-2021-annotating, caselli-etal-2021-dalc-fixed, nurce2022detectingabusivealbanian, Trajano_Bordini_Vieira_2024, rawat-etal-2023-modelling, singh_2024_mimic, Yuan_2025_generalizing} & 25 \\ \spacerow

Heuristics & \small\citet{ElSherief_Nilizadeh_Nguyen_Vigna_Belding_2018, Salminen_Almerekhi_Milenković_Jung_An_Kwak_Jansen_2018, kennedy_2020_constructing, albanyan-etal-2023-counterhate, Sarker_2023_automated, kirk-etal-2023-semeval} & 7 \\ \spacerow

Using all available data      & \small\citet{pavlopoulos-etal-2017-deep, ljubesic-etal-2018-datasets, Shekhar_Pranjić_Pollak_Pelicon_Purver_2020, assenmacher2021textttrpmod} & 7 \\ \spacerow

Geolocation     & \small\citet{mathur-etal-2018-offend-fixed, alvarez2018overview, caselli-etal-2021-dalc-fixed, castillo-lopez-etal-2023-analyzing, vasquez-etal-2023-homo} & 5 \\ \spacerow

Other     & \small\citet{mubarak-etal-2017-abusive, Wulczyn_2017_ex, Pelle_2017_offensive, de-gibert-etal-2018-hate, Founta_Djouvas_Chatzakou_Leontiadis_Blackburn_Stringhini_Vakali_Sirivianos_Kourtellis_2018, sprugnoli-etal-2018-creating, moon-etal-2020-beep, coltekin-2020-corpus, mollas_ethos_2022, Kennedy_2022_gab, Madhu_2023_detecting, ng-etal-2024-sghatecheck} & 12 \\ \spacerow

\textit{not reported}      & \small\citet{zueva-etal-2020-reducing, shekhar-etal-2022-coral-fixed, Trajano_Bordini_Vieira_2024, Sreelakshmi_2024_detection} & 4 \\ \spacerow

\bottomrule
\end{tabularx}%
\caption{Breakdown of datasets by collection methods.}
\label{tab:apx_method_breakdown}
\end{table}

\begin{table}[ht]
\begin{tabularx}{\textwidth}{@{}>{\centering\arraybackslash}m{3.5cm}Xc@{}}
\toprule
\textbf{Task formulation}    & \multicolumn{1}{c}{\textbf{Datasets}} & \multicolumn{1}{c}{\textbf{Count}} \\
\midrule
Binary classification & \small\citet{gao-huang-2017-detecting, Golbeck_2017_large, Wulczyn_2017_ex, ross_2016_measuring, pavlopoulos-etal-2017-deep, Alfina_2017_indonesian, Alakrot_2018_dataset, ljubesic-etal-2018-datasets, ElSherief_Nilizadeh_Nguyen_Vigna_Belding_2018, bohra-etal-2018-dataset, alvarez2018overview, qian-etal-2019-benchmark, suryawanshi-etal-2020-multimodal, pavlopoulos-etal-2020-toxicity, raman_2020_stress, Romim_2021_hate_bengali, assenmacher2021textttrpmod, saitov-derczynski-2021-abusive, kirk-etal-2022-hatemoji, Sarker_2023_automated, korre-etal-2023-harmful, park-etal-2023-uncovering, Das_Raj_Saha_Mathew_Gupta_Mukherjee_2023, Madhu_2023_detecting, goldzycher-etal-2024-improving, cignarella-etal-2024-queereotypes, ilevbare-etal-2024-ekohate, dementieva-etal-2024-toxicity, Ferreira_2024_towards, lee-etal-2024-exploring-cross, Sreelakshmi_2024_detection}  &  34  \\ \spacerow
Multi-class classification & \small\citet{waseem-hovy-2016-hateful-fixed, waseem-2016-racist-fixed, mubarak-etal-2017-abusive, Davidson_Warmsley_Macy_Weber_2017, jha-mamidi-2017-compliment-fixed, de-gibert-etal-2018-hate, Rezvan_2018_quality, mathur-etal-2018-offend-fixed, Ibrohim_2018_preliminaries, mulki-etal-2019-l, caselli-etal-2020-feel-fixed, Wijesiriwardene_2020_alone, kurrek-etal-2020-towards, Gomez_2020_exploring, pitenis-etal-2020-offensive, moon-etal-2020-beep, leite-etal-2020-toxic, grimminger-klinger-2021-hate, toraman-etal-2022-large, castillo-lopez-etal-2023-analyzing, rawat-etal-2023-modelling, Yuan_2025_generalizing} & 24 \\ \spacerow

Multi-label classification & \small\citet{Founta_Djouvas_Chatzakou_Leontiadis_Blackburn_Stringhini_Vakali_Sirivianos_Kourtellis_2018, ibrohim-budi-2019-multi, shekhar-etal-2022-coral-fixed, Kennedy_2022_gab} & 4 \\ \spacerow

Hierarchical & \small\citet{Bretschneider2017DetectingOS, pavlopoulos-etal-2017-deep, Pelle_2017_offensive, Fersini_2018_overview, Salminen_Almerekhi_Milenković_Jung_An_Kwak_Jansen_2018, WiegandSiegelRuppenhofer2019, kumar-etal-2018-aggression, sanguinetti-etal-2018-italian, Albadi_2018_are, bosco2018overview, sprugnoli-etal-2018-creating, zampieri-etal-2019-predicting, chung-etal-2019-conan, basile-etal-2019-semeval, Mandl_2019_hasoc, Ptaszynski_Pieciukiewicz_Dybala_2019, fortuna-etal-2019-hierarchically, Shekhar_Pranjić_Pollak_Pelicon_Purver_2020, sigurbergsson-derczynski-2020-offensive, bhardwaj2020hostilitydetectiondatasethindi, coltekin-2020-corpus, rizwan-etal-2020-hate, zeinert-etal-2021-annotating, caselli-etal-2021-dalc-fixed, cercas-curry-etal-2021-convabuse, vidgen-etal-2021-learning, rottger-etal-2021-hatecheck, Mathew_Saha_Yimam_Biemann_Goyal_Mukherjee_2021, vidgen-etal-2021-introducing-fixed, mollas_ethos_2022, nurce2022detectingabusivealbanian, JIANG2022swsr, kirk-etal-2022-hatemoji, albanyan-etal-2023-counterhate, Trajano_Bordini_Vieira_2024, kirk-etal-2023-semeval, vasquez-etal-2023-homo, raihan-etal-2023-offensive, Saeed_2023_detection, vargas-etal-2024-hausahate, singh_2024_mimic, ng-etal-2024-sghatecheck} & 54 \\ \spacerow

Parallel  & \small\citet{ousidhoum-etal-2019-multilingual, Fersini_Nozza_Rosso_2020, mulki-ghanem-2021-mi, Steffen_2023_codes, cignarella-etal-2024-queereotypes} & 7 \\ \spacerow

Other  & \small\citet{pamungkas-etal-2020-really-fixed, zueva-etal-2020-reducing, Samory_Sen_Kohne_Flöck_Wagner_2021, pavlopoulos-etal-2021-semeval, ollagnier-etal-2022-cyberagressionado, saker2023toxispanseexplainabletoxicitydetection} & 6 \\ \spacerow

\textit{not reported}      & \small\citet{zueva-etal-2020-reducing, shekhar-etal-2022-coral-fixed, Trajano_Bordini_Vieira_2024, Sreelakshmi_2024_detection} & 4 \\ \spacerow

\bottomrule
\end{tabularx}%
\caption{Breakdown of datasets by task types.}
\label{tab:apx_task_breakdown}
\end{table}

\begin{table}[ht]
\begin{tabularx}{\textwidth}{@{}>{\centering\arraybackslash}m{3.5cm}Xc@{}}
\toprule
\textbf{Number of annotators}    & \multicolumn{1}{c}{\textbf{Datasets}} & \multicolumn{1}{c}{\textbf{Count}} \\
\midrule
Involving single annotator (partially or fully) & \small\citet{gao-huang-2017-detecting, ljubesic-etal-2018-datasets, Salminen_Almerekhi_Milenković_Jung_An_Kwak_Jansen_2018, WiegandSiegelRuppenhofer2019, ibrohim-budi-2019-multi, pamungkas-etal-2020-really-fixed, suryawanshi-etal-2020-multimodal, coltekin-2020-corpus, caselli-etal-2021-dalc-fixed, vidgen-etal-2021-learning, grimminger-klinger-2021-hate, ollagnier-etal-2022-cyberagressionado, goldzycher-etal-2024-improving, Ferreira_2024_towards}  &  14  \\ \spacerow

Multiple, subset & \small\citet{ross_2016_measuring, Alfina_2017_indonesian, Ibrohim_2018_preliminaries, bosco2018overview, ibrohim-budi-2019-multi, Ptaszynski_Pieciukiewicz_Dybala_2019, fortuna-etal-2019-hierarchically, pamungkas-etal-2020-really-fixed, suryawanshi-etal-2020-multimodal, kurrek-etal-2020-towards, vidgen-etal-2020-detecting, leite-etal-2020-toxic, Romim_2021_hate_bengali, zeinert-etal-2021-annotating, cercas-curry-etal-2021-convabuse, vidgen-etal-2021-learning, grimminger-klinger-2021-hate, rottger-etal-2021-hatecheck, shekhar-etal-2022-coral-fixed, toraman-etal-2022-large, kirk-etal-2022-hatemoji, demus-etal-2022-comprehensive-fixed, kirk-etal-2023-semeval, Steffen_2023_codes, raihan-etal-2023-offensive, Das_Raj_Saha_Mathew_Gupta_Mukherjee_2023, Madhu_2023_detecting} & 29 \\ \spacerow

Multiple, full set & \small\citet{Golbeck_2017_large, Bretschneider2017DetectingOS, pavlopoulos-etal-2017-deep, Pelle_2017_offensive, Alakrot_2018_dataset, ljubesic-etal-2018-datasets, Fersini_2018_overview, Rezvan_2018_quality, mathur-etal-2018-offend-fixed, bohra-etal-2018-dataset, sprugnoli-etal-2018-creating, alvarez2018overview, mulki-etal-2019-l, chung-etal-2019-conan, caselli-etal-2020-feel-fixed, Wijesiriwardene_2020_alone, pitenis-etal-2020-offensive, rizwan-etal-2020-hate, mulki-ghanem-2021-mi, caselli-etal-2021-dalc-fixed, pavlopoulos-etal-2021-semeval, fanton-etal-2021-human-fixed, vidgen-etal-2021-introducing-fixed, saitov-derczynski-2021-abusive, nurce2022detectingabusivealbanian, JIANG2022swsr, kirk-etal-2022-hatemoji, Kennedy_2022_gab, albanyan-etal-2023-counterhate, saker2023toxispanseexplainabletoxicitydetection, Sarker_2023_automated, park-etal-2023-uncovering, castillo-lopez-etal-2023-analyzing, rawat-etal-2023-modelling, vasquez-etal-2023-homo, Saeed_2023_detection, cignarella-etal-2024-queereotypes, ilevbare-etal-2024-ekohate, vargas-etal-2024-hausahate, singh_2024_mimic, lee-etal-2024-exploring-cross, Sreelakshmi_2024_detection} & 47 \\ \spacerow

Involving crowdsourcing & \small\citet{mubarak-etal-2017-abusive, Davidson_Warmsley_Macy_Weber_2017, Wulczyn_2017_ex, Albadi_2018_are, Fersini_2018_overview, Founta_Djouvas_Chatzakou_Leontiadis_Blackburn_Stringhini_Vakali_Sirivianos_Kourtellis_2018, ousidhoum-etal-2019-multilingual, zampieri-etal-2019-predicting, Borkan_2019_muanced, qian-etal-2019-benchmark, basile-etal-2019-semeval, kennedy_2020_constructing, Gomez_2020_exploring, pavlopoulos-etal-2020-toxicity, Samory_Sen_Kohne_Flöck_Wagner_2021, Mathew_Saha_Yimam_Biemann_Goyal_Mukherjee_2021, mollas_ethos_2022, assenmacher2021textttrpmod, kumar-etal-2018-aggression, moon-etal-2020-beep, korre-etal-2023-harmful, Yuan_2025_generalizing} & 29 \\ \spacerow

\textit{not reported or unclear} & \small\citet{waseem-hovy-2016-hateful-fixed, jha-mamidi-2017-compliment-fixed, ljubesic-etal-2018-datasets, de-gibert-etal-2018-hate, ElSherief_Nilizadeh_Nguyen_Vigna_Belding_2018, bosco2018overview, Mandl_2019_hasoc, Shekhar_Pranjić_Pollak_Pelicon_Purver_2020, sigurbergsson-derczynski-2020-offensive, bhardwaj2020hostilitydetectiondatasethindi, Fersini_Nozza_Rosso_2020, zueva-etal-2020-reducing, raman_2020_stress, Trajano_Bordini_Vieira_2024, seo-etal-2024-kocommongen, ng-etal-2024-sghatecheck, dementieva-etal-2024-toxicity} & 23 \\ \spacerow

\bottomrule
\end{tabularx}%
\caption{Breakdown of datasets by numbers of annotators.}
\label{tab:apx_anno_breakdown}
\end{table}

\begin{table}[ht]
\begin{tabularx}{\textwidth}{@{}>{\centering\arraybackslash}m{3.5cm}Xc@{}}
\toprule
\textbf{Reported Demographics}    & \multicolumn{1}{c}{\textbf{Datasets}} & \multicolumn{1}{c}{\textbf{Count}} \\
\midrule
Age & \footnotesize\citet{ross_2016_measuring, Alfina_2017_indonesian, Alakrot_2018_dataset, Founta_Djouvas_Chatzakou_Leontiadis_Blackburn_Stringhini_Vakali_Sirivianos_Kourtellis_2018, chung-etal-2019-conan, ibrohim-budi-2019-multi, sigurbergsson-derczynski-2020-offensive, kurrek-etal-2020-towards, vidgen-etal-2020-detecting, leite-etal-2020-toxic, zeinert-etal-2021-annotating, caselli-etal-2021-dalc-fixed, cercas-curry-etal-2021-convabuse, vidgen-etal-2021-learning, grimminger-klinger-2021-hate, rottger-etal-2021-hatecheck, vidgen-etal-2021-introducing-fixed, assenmacher2021textttrpmod, saitov-derczynski-2021-abusive, nurce2022detectingabusivealbanian, toraman-etal-2022-large, kirk-etal-2022-hatemoji, kirk-etal-2023-semeval, vasquez-etal-2023-homo, raihan-etal-2023-offensive, goldzycher-etal-2024-improving, ng-etal-2024-sghatecheck, lee-etal-2024-exploring-cross}  &  33  \\ \spacerow
Gender & \footnotesize\citet{ross_2016_measuring, Alfina_2017_indonesian, Founta_Djouvas_Chatzakou_Leontiadis_Blackburn_Stringhini_Vakali_Sirivianos_Kourtellis_2018, chung-etal-2019-conan, ibrohim-budi-2019-multi, sigurbergsson-derczynski-2020-offensive, suryawanshi-etal-2020-multimodal, kurrek-etal-2020-towards, vidgen-etal-2020-detecting, leite-etal-2020-toxic, mulki-ghanem-2021-mi, zeinert-etal-2021-annotating, caselli-etal-2021-dalc-fixed, cercas-curry-etal-2021-convabuse, vidgen-etal-2021-learning, grimminger-klinger-2021-hate, rottger-etal-2021-hatecheck, vidgen-etal-2021-introducing-fixed, assenmacher2021textttrpmod, saitov-derczynski-2021-abusive, nurce2022detectingabusivealbanian, JIANG2022swsr, kirk-etal-2022-hatemoji, saker2023toxispanseexplainabletoxicitydetection, vasquez-etal-2023-homo, raihan-etal-2023-offensive, goldzycher-etal-2024-improving, ilevbare-etal-2024-ekohate, ng-etal-2024-sghatecheck, lee-etal-2024-exploring-cross} & 33 \\ \spacerow
Language & \footnotesize\citet{gao-huang-2017-detecting, Albadi_2018_are, Rezvan_2018_quality, WiegandSiegelRuppenhofer2019, ousidhoum-etal-2019-multilingual, chung-etal-2019-conan, sigurbergsson-derczynski-2020-offensive, vidgen-etal-2020-detecting, coltekin-2020-corpus, mulki-ghanem-2021-mi, zeinert-etal-2021-annotating, caselli-etal-2021-dalc-fixed, cercas-curry-etal-2021-convabuse, vidgen-etal-2021-learning, grimminger-klinger-2021-hate, rottger-etal-2021-hatecheck, saitov-derczynski-2021-abusive, nurce2022detectingabusivealbanian, kirk-etal-2022-hatemoji, castillo-lopez-etal-2023-analyzing, kirk-etal-2023-semeval, vasquez-etal-2023-homo, raihan-etal-2023-offensive, goldzycher-etal-2024-improving, vargas-etal-2024-hausahate, ng-etal-2024-sghatecheck} & 32 \\ 
\spacerow

Education & \footnotesize\citet{Founta_Djouvas_Chatzakou_Leontiadis_Blackburn_Stringhini_Vakali_Sirivianos_Kourtellis_2018, chung-etal-2019-conan, ibrohim-budi-2019-multi, vidgen-etal-2020-detecting, Romim_2021_hate_bengali, caselli-etal-2021-dalc-fixed, cercas-curry-etal-2021-convabuse, vidgen-etal-2021-learning, grimminger-klinger-2021-hate, rottger-etal-2021-hatecheck, toraman-etal-2022-large, kirk-etal-2022-hatemoji, rawat-etal-2023-modelling, kirk-etal-2023-semeval, vasquez-etal-2023-homo, raihan-etal-2023-offensive, Das_Raj_Saha_Mathew_Gupta_Mukherjee_2023, goldzycher-etal-2024-improving, ilevbare-etal-2024-ekohate, vargas-etal-2024-hausahate, ng-etal-2024-sghatecheck, lee-etal-2024-exploring-cross} & 27 \\ \spacerow

Location (nationality, country of origin, IP) & \footnotesize\citet{mubarak-etal-2017-abusive, Albadi_2018_are, Alakrot_2018_dataset, Founta_Djouvas_Chatzakou_Leontiadis_Blackburn_Stringhini_Vakali_Sirivianos_Kourtellis_2018, mulki-etal-2019-l, vidgen-etal-2020-detecting, vidgen-etal-2021-learning, rottger-etal-2021-hatecheck, vidgen-etal-2021-introducing-fixed, assenmacher2021textttrpmod, nurce2022detectingabusivealbanian, kirk-etal-2022-hatemoji, castillo-lopez-etal-2023-analyzing, kirk-etal-2023-semeval, vasquez-etal-2023-homo, vargas-etal-2024-hausahate} & 18 \\ \spacerow

Race and ethnicity & \footnotesize\citet{Alfina_2017_indonesian, ibrohim-budi-2019-multi, sigurbergsson-derczynski-2020-offensive, kurrek-etal-2020-towards, leite-etal-2020-toxic, zeinert-etal-2021-annotating, caselli-etal-2021-dalc-fixed, cercas-curry-etal-2021-convabuse, vidgen-etal-2021-learning, rottger-etal-2021-hatecheck, vidgen-etal-2021-introducing-fixed, kirk-etal-2022-hatemoji, kirk-etal-2023-semeval, lee-etal-2024-exploring-cross} & 16 \\ \spacerow


\textit{not reported} & \scriptsize\citet{waseem-hovy-2016-hateful-fixed, waseem-2016-racist-fixed, mubarak-etal-2017-abusive, Davidson_Warmsley_Macy_Weber_2017, jha-mamidi-2017-compliment-fixed, Golbeck_2017_large, Wulczyn_2017_ex, Bretschneider2017DetectingOS, pavlopoulos-etal-2017-deep, Pelle_2017_offensive, ljubesic-etal-2018-datasets, de-gibert-etal-2018-hate, Fersini_2018_overview, ElSherief_Nilizadeh_Nguyen_Vigna_Belding_2018, Salminen_Almerekhi_Milenković_Jung_An_Kwak_Jansen_2018, kumar-etal-2018-aggression, mathur-etal-2018-offend-fixed, bohra-etal-2018-dataset, Ibrohim_2018_preliminaries, sanguinetti-etal-2018-italian, bosco2018overview, sprugnoli-etal-2018-creating, alvarez2018overview, zampieri-etal-2019-predicting, Borkan_2019_muanced, qian-etal-2019-benchmark, basile-etal-2019-semeval, Mandl_2019_hasoc, Ptaszynski_Pieciukiewicz_Dybala_2019, fortuna-etal-2019-hierarchically, Shekhar_Pranjić_Pollak_Pelicon_Purver_2020, kennedy_2020_constructing, caselli-etal-2020-feel-fixed, pamungkas-etal-2020-really-fixed, Wijesiriwardene_2020_alone, Gomez_2020_exploring, pavlopoulos-etal-2020-toxicity, pitenis-etal-2020-offensive, bhardwaj2020hostilitydetectiondatasethindi, moon-etal-2020-beep, Fersini_Nozza_Rosso_2020, zueva-etal-2020-reducing, rizwan-etal-2020-hate, raman_2020_stress, Samory_Sen_Kohne_Flöck_Wagner_2021, pavlopoulos-etal-2021-semeval, fanton-etal-2021-human-fixed, Mathew_Saha_Yimam_Biemann_Goyal_Mukherjee_2021, mollas_ethos_2022, shekhar-etal-2022-coral-fixed, ollagnier-etal-2022-cyberagressionado, demus-etal-2022-comprehensive-fixed, albanyan-etal-2023-counterhate, Sarker_2023_automated, Trajano_Bordini_Vieira_2024, korre-etal-2023-harmful, park-etal-2023-uncovering, Madhu_2023_detecting, cignarella-etal-2024-queereotypes, seo-etal-2024-kocommongen, dementieva-etal-2024-toxicity, singh_2024_mimic, Ferreira_2024_towards, Yuan_2025_generalizing} & 78 \\ \spacerow

\bottomrule
\end{tabularx}%
\caption{Examples of annotator demographics and datasets that report them.}
\label{tab:apx_anno_demo_breakdown}
\end{table}

\begin{table}[ht]
\begin{tabularx}{\textwidth}{@{}>{\centering\arraybackslash}m{3.5cm}Xc@{}}
\toprule
\textbf{Methods to resolve disagreements}    & \multicolumn{1}{c}{\textbf{Datasets}} & \multicolumn{1}{c}{\textbf{Count}} \\
\midrule
Majority vote & \small\citet{Samory_Sen_Kohne_Flöck_Wagner_2021, qian-etal-2019-benchmark, fortuna-etal-2019-hierarchically, Rezvan_2018_quality, Wijesiriwardene_2020_alone, waseem-2016-racist-fixed, Davidson_Warmsley_Macy_Weber_2017, moon-etal-2020-beep, shekhar-etal-2022-coral-fixed, Alakrot_2018_dataset, pavlopoulos-etal-2017-deep, Sreelakshmi_2024_detection, Saeed_2023_detection, demus-etal-2022-comprehensive-fixed, mathur-etal-2018-offend-fixed, Wulczyn_2017_ex, lee-etal-2024-exploring-cross, Gomez_2020_exploring, korre-etal-2023-harmful, Romim_2021_hate_bengali, Mathew_Saha_Yimam_Biemann_Goyal_Mukherjee_2021, vargas-etal-2024-hausahate, vasquez-etal-2023-homo, caselli-etal-2020-feel-fixed, Kennedy_2022_gab, mulki-etal-2019-l, Founta_Djouvas_Chatzakou_Leontiadis_Blackburn_Stringhini_Vakali_Sirivianos_Kourtellis_2018, toraman-etal-2022-large, mulki-ghanem-2021-mi, ibrohim-budi-2019-multi, ousidhoum-etal-2019-multilingual, suryawanshi-etal-2020-multimodal, Pelle_2017_offensive, pitenis-etal-2020-offensive, Trajano_Bordini_Vieira_2024, Fersini_2018_overview, ElSherief_Nilizadeh_Nguyen_Vigna_Belding_2018, zampieri-etal-2019-predicting, basile-etal-2019-semeval, pavlopoulos-etal-2021-semeval}  &  48  \\ \spacerow

Additional annotators & \small\citet{Golbeck_2017_large, Fersini_2018_overview, mathur-etal-2018-offend-fixed, sanguinetti-etal-2018-italian, zampieri-etal-2019-predicting, basile-etal-2019-semeval, Ptaszynski_Pieciukiewicz_Dybala_2019, pamungkas-etal-2020-really-fixed, kurrek-etal-2020-towards, vidgen-etal-2020-detecting, coltekin-2020-corpus, vidgen-etal-2021-introducing-fixed, toraman-etal-2022-large, kirk-etal-2022-hatemoji, castillo-lopez-etal-2023-analyzing, rawat-etal-2023-modelling, kirk-etal-2023-semeval, raihan-etal-2023-offensive, Das_Raj_Saha_Mathew_Gupta_Mukherjee_2023, Madhu_2023_detecting, goldzycher-etal-2024-improving, cignarella-etal-2024-queereotypes} & 27 \\ \spacerow

Moderation meeting & \small\citet{Salminen_Almerekhi_Milenković_Jung_An_Kwak_Jansen_2018, zeinert-etal-2021-annotating, caselli-etal-2021-dalc-fixed, albanyan-etal-2023-counterhate, saker2023toxispanseexplainabletoxicitydetection, Sarker_2023_automated, ilevbare-etal-2024-ekohate, Ferreira_2024_towards} & 8 \\ \spacerow

Other & \small\citet{gao-huang-2017-detecting, Bretschneider2017DetectingOS, Albadi_2018_are, Alakrot_2018_dataset, pavlopoulos-etal-2020-toxicity, leite-etal-2020-toxic, assenmacher2021textttrpmod, Trajano_Bordini_Vieira_2024, vasquez-etal-2023-homo, Yuan_2025_generalizing} & 12 \\ \spacerow

Discarded & \small\citet{Davidson_Warmsley_Macy_Weber_2017, Alfina_2017_indonesian, Albadi_2018_are, Ibrohim_2018_preliminaries, mulki-etal-2019-l, ibrohim-budi-2019-multi, rizwan-etal-2020-hate, mulki-ghanem-2021-mi, Mathew_Saha_Yimam_Biemann_Goyal_Mukherjee_2021} & 9 \\ 
\spacerow

\textit{not applicable} & \small\citet{ross_2016_measuring, ljubesic-etal-2018-datasets, WiegandSiegelRuppenhofer2019, chung-etal-2019-conan, Shekhar_Pranjić_Pollak_Pelicon_Purver_2020, kennedy_2020_constructing, vidgen-etal-2021-learning, grimminger-klinger-2021-hate, rottger-etal-2021-hatecheck, ollagnier-etal-2022-cyberagressionado, ng-etal-2024-sghatecheck} & 16 \\ \spacerow

\textit{not reported} & \small\citet{waseem-hovy-2016-hateful-fixed, mubarak-etal-2017-abusive, jha-mamidi-2017-compliment-fixed, de-gibert-etal-2018-hate, kumar-etal-2018-aggression, bohra-etal-2018-dataset, bosco2018overview, sprugnoli-etal-2018-creating, alvarez2018overview, Borkan_2019_muanced, Mandl_2019_hasoc, sigurbergsson-derczynski-2020-offensive, bhardwaj2020hostilitydetectiondatasethindi, Fersini_Nozza_Rosso_2020, zueva-etal-2020-reducing, raman_2020_stress, fanton-etal-2021-human-fixed, mollas_ethos_2022, saitov-derczynski-2021-abusive, nurce2022detectingabusivealbanian, JIANG2022swsr, park-etal-2023-uncovering, seo-etal-2024-kocommongen, dementieva-etal-2024-toxicity, singh_2024_mimic} & 31 \\ \spacerow

\bottomrule
\end{tabularx}%
\caption{Breakdown of datasets by label aggregation strategies.}
\label{tab:apx_aggre_breakdown}
\end{table}

\begin{table}[ht]
\begin{tabularx}{\textwidth}{@{}>{\centering\arraybackslash}m{3.5cm}Xc@{}}
\toprule
\textbf{Methods to resolve disagreements}    & \multicolumn{1}{c}{\textbf{Datasets}} & \multicolumn{1}{c}{\textbf{Count}} \\
\midrule

Metrics-based selection (crowdsourcing) & \small\citet{ElSherief_Nilizadeh_Nguyen_Vigna_Belding_2018, ousidhoum-etal-2019-multilingual, qian-etal-2019-benchmark, Samory_Sen_Kohne_Flöck_Wagner_2021, Mathew_Saha_Yimam_Biemann_Goyal_Mukherjee_2021, assenmacher2021textttrpmod, Yuan_2025_generalizing} & 10 \\ \spacerow

Training & \small\citet{Golbeck_2017_large, kurrek-etal-2020-towards, vidgen-etal-2020-detecting, vidgen-etal-2021-learning, vidgen-etal-2021-introducing-fixed, shekhar-etal-2022-coral-fixed, Kennedy_2022_gab, demus-etal-2022-comprehensive-fixed, Trajano_Bordini_Vieira_2024, vasquez-etal-2023-homo, dementieva-etal-2024-toxicity} & 13 \\ 
\spacerow

Moderation meetings only to resolve disagreements & \small\citet{gao-huang-2017-detecting, Golbeck_2017_large, caselli-etal-2020-feel-fixed, kurrek-etal-2020-towards, zeinert-etal-2021-annotating, caselli-etal-2021-dalc-fixed, cercas-curry-etal-2021-convabuse, kirk-etal-2022-hatemoji, ollagnier-etal-2022-cyberagressionado, demus-etal-2022-comprehensive-fixed, vasquez-etal-2023-homo, Das_Raj_Saha_Mathew_Gupta_Mukherjee_2023} & 12 \\ \spacerow

Moderation meetings to refine guidelines & \small\citet{kumar-etal-2018-aggression, suryawanshi-etal-2020-multimodal, JIANG2022swsr, kirk-etal-2023-semeval, park-etal-2023-uncovering, raihan-etal-2023-offensive, cignarella-etal-2024-queereotypes} & 10 \\ \spacerow

Tests (During onboarding or hidden during annotation) & \small\citet{Wulczyn_2017_ex, ElSherief_Nilizadeh_Nguyen_Vigna_Belding_2018, Albadi_2018_are, zampieri-etal-2019-predicting, basile-etal-2019-semeval, Samory_Sen_Kohne_Flöck_Wagner_2021, mollas_ethos_2022, assenmacher2021textttrpmod, korre-etal-2023-harmful, kirk-etal-2023-semeval, lee-etal-2024-exploring-cross} & 12 \\ \spacerow

Validation by outside annotators & \small\citet{waseem-hovy-2016-hateful-fixed, jha-mamidi-2017-compliment-fixed, Salminen_Almerekhi_Milenković_Jung_An_Kwak_Jansen_2018, Romim_2021_hate_bengali, vidgen-etal-2021-learning, rottger-etal-2021-hatecheck, goldzycher-etal-2024-improving, Ptaszynski_Pieciukiewicz_Dybala_2019, ilevbare-etal-2024-ekohate, dementieva-etal-2024-toxicity}  &  10  \\ \spacerow

\textit{not reported or unclear} & \small\citet{waseem-2016-racist-fixed, mubarak-etal-2017-abusive, Davidson_Warmsley_Macy_Weber_2017, ross_2016_measuring, Bretschneider2017DetectingOS, Alfina_2017_indonesian, Pelle_2017_offensive, Alakrot_2018_dataset, ljubesic-etal-2018-datasets, Founta_Djouvas_Chatzakou_Leontiadis_Blackburn_Stringhini_Vakali_Sirivianos_Kourtellis_2018, Rezvan_2018_quality, mathur-etal-2018-offend-fixed, bohra-etal-2018-dataset, Ibrohim_2018_preliminaries, bosco2018overview, sprugnoli-etal-2018-creating, alvarez2018overview, mulki-etal-2019-l, Borkan_2019_muanced, chung-etal-2019-conan, basile-etal-2019-semeval, Mandl_2019_hasoc, ibrohim-budi-2019-multi, fortuna-etal-2019-hierarchically, Shekhar_Pranjić_Pollak_Pelicon_Purver_2020, sigurbergsson-derczynski-2020-offensive, pamungkas-etal-2020-really-fixed, Wijesiriwardene_2020_alone, pavlopoulos-etal-2020-toxicity, pitenis-etal-2020-offensive, bhardwaj2020hostilitydetectiondatasethindi, moon-etal-2020-beep, Fersini_Nozza_Rosso_2020, leite-etal-2020-toxic, zueva-etal-2020-reducing, coltekin-2020-corpus, rizwan-etal-2020-hate, raman_2020_stress, mulki-ghanem-2021-mi, pavlopoulos-etal-2021-semeval, fanton-etal-2021-human-fixed, saitov-derczynski-2021-abusive, nurce2022detectingabusivealbanian, toraman-etal-2022-large, kirk-etal-2022-hatemoji, albanyan-etal-2023-counterhate, saker2023toxispanseexplainabletoxicitydetection, Sarker_2023_automated, castillo-lopez-etal-2023-analyzing, rawat-etal-2023-modelling, Steffen_2023_codes, Saeed_2023_detection, Madhu_2023_detecting, cignarella-etal-2024-queereotypes, seo-etal-2024-kocommongen, vargas-etal-2024-hausahate, ng-etal-2024-sghatecheck, singh_2024_mimic, Ferreira_2024_towards, Sreelakshmi_2024_detection} & 70 \\ \spacerow

\bottomrule
\end{tabularx}%
\caption{Breakdown of datasets by quality assurance steps.}
\label{tab:apx_qa_breakdown}
\end{table}

\end{document}